\documentclass[letterpaper]{article}
\usepackage[top=1.00in, bottom=1.25in, left=1.25in, right=1.25in]{geometry}

\usepackage{microtype}
\usepackage{amsmath}
\usepackage{graphicx}
\usepackage{amsfonts}
\usepackage{algorithm}
\usepackage{algorithmic}
\usepackage{xcolor}
\usepackage{multirow}
\usepackage{booktabs}
\usepackage{natbib}
\usepackage{authblk}
\usepackage{mathtools}
\usepackage{caption}
\newcommand{\argmax}{\operatornamewithlimits{\arg \max}}

\newcommand{\bh}{{\mathbf{h}}}
\newcommand{\bH}{{\mathbf{H}}}
\newcommand{\bx}{{\mathbf{x}}}
\newcommand{\bz}{{\mathbf{z}}}

\newcommand{\bmu}{{\boldsymbol \mu}}
\newcommand{\bpsi}{{\boldsymbol \psi}}
\newcommand{\bsigma}{{\boldsymbol \sigma}}
\newcommand{\btheta}{{\boldsymbol \theta}}

\newcommand{\calE}{\mathcal{E}}

\newcommand{\calX}{\mathcal{X}}
\newcommand{\calY}{\mathcal{Y}}

\newcommand{\bbE}{\mathbb{E}}
\newcommand{\bbR}{\mathbb{R}}

\newcommand{\hbh}{{\mathbf{\hat{h}}}}

\definecolor{orange}{RGB}{255,127,0}

\begin{document}

\title{MxML: Mixture of Meta-Learners\\for Few-Shot Classification}

\author[1]{Minseop Park}
\author[2]{Jungtaek Kim}
\author[1]{Saehoon Kim}
\author[3]{Yanbin Liu}
\author[2]{Seungjin Choi}
\affil[1]{\normalsize AITRICS, Seoul, Republic of Korea}
\affil[2]{\normalsize Pohang University of Science and Technology, Pohang, Republic of Korea}
\affil[3]{\normalsize University of Technology Sydney, Sydney, Australia}
\affil[ ]{\small \texttt{\{mike\_seop, shkim\}@aitrics.com}, \texttt{\{jtkim,seungjin\}@postech.ac.kr}, \texttt{csyanbin@gmail.com}}

\maketitle

\begin{abstract}
A meta-model is trained on a distribution of similar tasks such that it learns an algorithm that can quickly adapt to a novel task with only a handful of labeled examples. Most of current meta-learning methods assume that the meta-training set consists of relevant tasks sampled from a single distribution. In practice, however, a new task is often out of the task distribution, yielding a performance degradation. One way to tackle this problem is to construct an ensemble of meta-learners such that each meta-learner is trained on different task distribution. In this paper we present a method for constructing a mixture of meta-learners
(MxML), where mixing parameters are determined by the weight prediction network (WPN) optimized to improve the few-shot classification performance. Experiments on various datasets demonstrate that MxML
significantly outperforms state-of-the-art meta-learners, or their na\"ive ensemble in the case of out-of-distribution as well as in-distribution tasks.
\end{abstract}


\section{Introduction\label{sec:intro}}

Deep neural networks, trained over a large-scale dataset with legitimate regularization techniques,
generalize to a novel instance with persistent performance, while they are highly over-parameterized.
Many attempts have been introduced to analyze their
generalization performance in notion of sharpness of local minima~\citep{KeskarNS2017iclr}, 
which describes the reason why deep networks can generalize even if the number of parameters
is larger than the number of training instances.
Yet, complex deep networks learned from few examples tend to be easily over-fitting
to the training set, which is hardly alleviated by a regularization
from Bayesian learning~\citep{FeiFeiL2003iccv,FeiFeiL2006ieeetpami,Salakhutdinov2012icmlworkshopuns}.

The primary interest of this paper is \emph{few-shot classification}:
the objective is to learn a mapping 
function that assigns each instance in a query set $Q$ into 
few-shot classes defined by a support set $S$,
which is composed of a set of few instances in classes.
Under this problem setting, meta-learning~\citep{SchmidhuberJ87phd,ThrunS98kluwer} generalizes to
a novel task by learning a series of tasks $\calE = \{E_i\}_{i = 1}^T$, where $E_i$ is the $i$th episode
that consists of a tuple of $Q$ and $S$, and $T$ is the number of 
training elements.
A common practice to train a meta-learner has a major limitation, 
where tasks in the phase of meta-training and meta-test
are sampled from the same dataset.
Similar to supervised learning in which training and test 
distributions are typically matched, meta-learning implicitly assumes that 
tasks from meta-training and meta-test share similar high-level concepts. 
Then, the learner performs poorly to a novel task 
that does not share common attributes of the tasks in meta-training.

\begin{figure}[t]
\begin{center}
\includegraphics[width=0.55\linewidth]{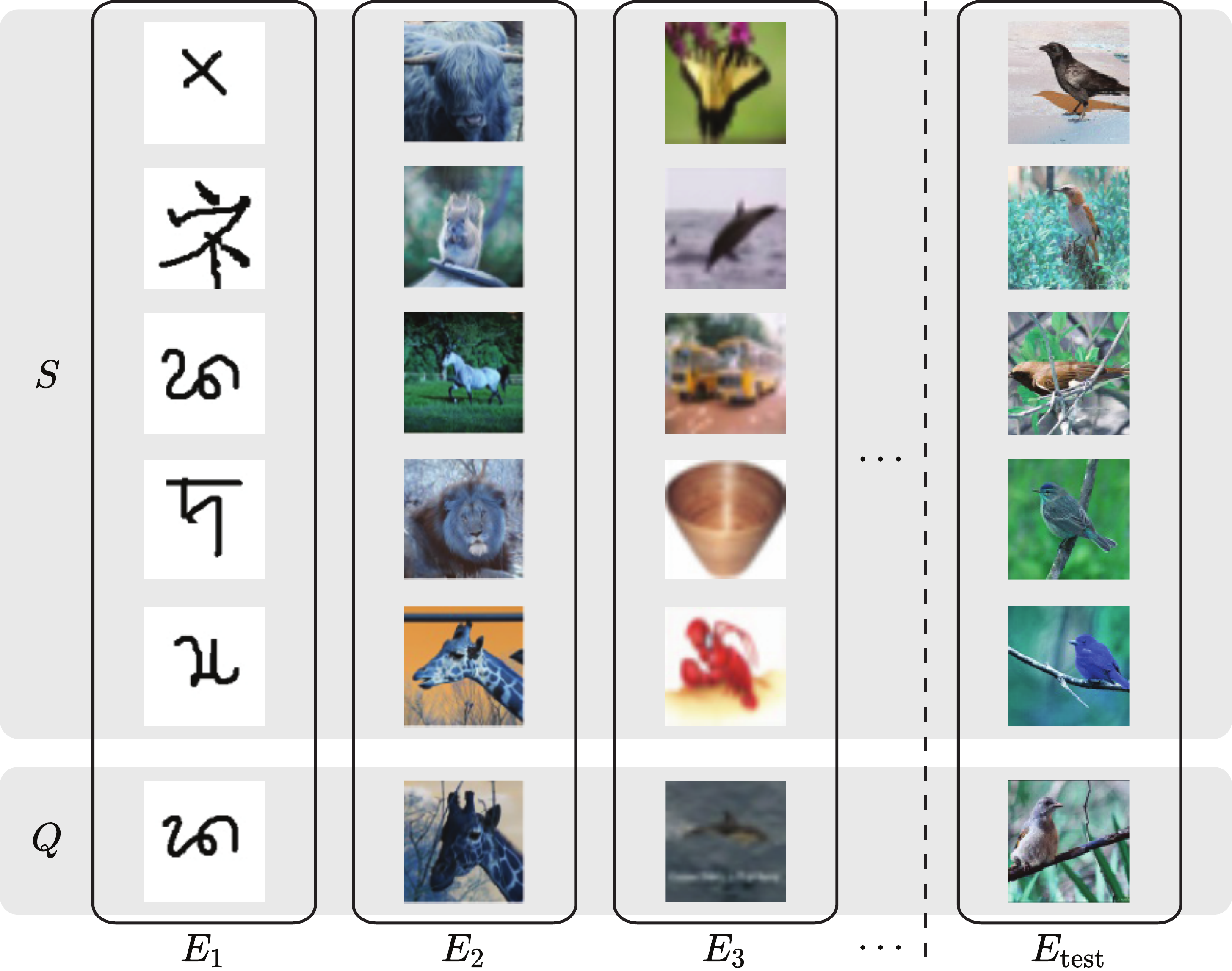}
\end{center}
\caption{
Illustration of solving out-of-distribution task in one-shot scenario. 
To obtain a model that is capable of learning a task from 
unseen visual domain, it is necessary to train the model 
with tasks from diverse distributions with meta-learning.
However, na\"ive training series of episodes $E_1,E_2,\ldots$ will 
degenerate the performance because of distractors.
}
\label{fig:concept}
\end{figure}

Rather than learning from a single dataset, we expect meta-learners 
to be trained over datasets from diverse visual domains, 
and perform robustly to a novel task 
by expanding their knowledge from the most similar ones to a novel task.
Yet, a na\"ive training from multiple sets 
does not perform well because the model considers many irrelevant tasks.
To alleviate the effects, an obvious approach is to retrieve 
similar ones of a novel task and to train a meta-learner 
from them.
This may perform well enough to the target task,
but we observe some limitations:
(i) learning an appropriate metric between datasets is 
quite challenging and (ii) a meta-learner is always 
trained from scratch whenever a new task is given.
Instead of selecting similar datasets~\citep{KimJ2017arxiv}, 
it is better to keep multiple meta-learners trained from each dataset
and determine how to aggregate them effectively.
This encourages us to build a mixture
of meta-learners where mixture coefficients are adaptively
determined whenever a novel task is given.

The major concern of building such model is that the model has to
determine the coefficient while it glimpses the target task, 
contrast to the regular supervised learning which has abundant 
validating examples to evaluate the base learners. To this end, 
we train the model to learn how to combine the meta-learners given an episode. 
More specifically, our mixture of meta-learners is 
established by putting more weight on the base-learner 
that is expected to perform well to the tasks in the test phase.
To evaluate the model given a small number of instances, 
we employ the weight prediction network (WPN) that predicts 
the performance of the base learner by observing 
their latent embeddings of given task. 
Since WPN determines the performance of the meta-learner 
based on its output,
it can be viewed as a similar idea of meta-recognition system~\citep{ScheirerWJ2012ieeetpami} that analyzes and predicts the recognition system. Hence, learning to evaluate meta-learners 
can be considered as two layers of meta-learning.

Our contribution is two-fold: 
(i) we point out a major limitation of the conventional approaches 
and propose MxML as a solution, such that the mixture coefficients on base
meta-learners are task-adaptively determined 
(ii) we observe that our model achieves the best performance among 
the state-of-the-art algorithms,
when the task is sampled from novel distribution (out-of-distribution)
as well as when the task shares the similar attributes with 
training tasks (in-distribution). 

\section{Background}

This section introduces a problem setting and noticeable works on meta-learning for few-shot classification.

\subsection{Problem Setting}

We follow the conventional definition of few-shot 
classification as in \citep{VinyalsO2016nips,SnellJ2017nips}.
The objective of few-shot classification
is to estimate a function $f(\bx ; \btheta)$ 
parameterized by $\btheta$ that maps an instance of 
a query set $Q$ into a label set $\calY$.
Specifically, 
the $N$-way $K$-shot classification is formally defined as
the task that assigns a query into one of $N$ 
classes in the support set composed of examples and their
associated labels: 
$S=\{(\bx_i, y_i)\}_{i=1}^{NK}$, 
where an example $\bx_i \in \calX$, the associated label $y_i \in \calY$, and $|\calY| = N$. Note that the number of 
examples with the same label is $K$ and $\calX$ represents
an input space. Similarly, $Q=\{(\bx_i, y_i)\}_{i=1}^{L}$,
where $L$ is the number of queries and the associated labels
are only given in the meta-training.

Meta-learning~\citep{SchmidhuberJ87phd,ThrunS98kluwer}
for few-shot classification introduces an episode,
a tuple of $Q$ and $S$ sampled from a dataset,
that is used to learn parameters of a model 
in an episodic training strategy~\citep{VinyalsO2016nips,SnellJ2017nips}.
It effectively prevents from over-fitting of a model
when it is solely trained by a single task.
In the subsequent section, we briefly summarize representative 
meta-learning methods for few-shot classification.

\subsection{Related Work}

We categorized previous works by the existence of adaptation to few labeled examples of a task in the test phase.

\subsubsection{Meta-Learning without Adaptation}

Learning appropriate metrics is a key step to solve few-shot learning. 
Along this direction, matching network~\citep{VinyalsO2016nips} 
proposes a differentiable nearest neighbor classifier that is learned to minimize the empirical risk computed in the meta-training phase. 
Given a set of few-shot classes, matching network learns a mapping function from a test instance into one of few-shot classes, which is formulated by bi-directional LSTM with attention mechanism.
Prototypical networks~\citep{SnellJ2017nips} simply learn a representative vector in each few-shot classes, instead of learning complex neural networks with attention mechanism. This is also trained by a series of episodes, where the prototype vectors are learned to enforce that they should be close in the same class. Moreover, its simple extension to learn covariance structures is also available at~\citep{FortS2017arxiv}.
Among early works on this direction, Siamese neural network~\citep{KochG2015icml} is used to learn the metric that preserves semantic similarities between instances.

\subsubsection{Meta-Learning with Adaptation}
Learning models that quickly adapt to few examples is
critical to solve few-shot learning. Along this direction,
model-agnostic meta-learning (MAML) explicitly 
trains a meta-learner such that few updates with labeled instances are enough to achieve high generalization performance on a new task~\citep{FinnC2017icml}.
The original implementation of MAML requires a second-order
derivative of parameters of deep neural network,
which is accelerated by a first-order approximation~\citep{NicholA2018arxiv}.
Similarly, \citet{RaviS2017iclr}
propose a meta-learning framework such that LSTM is trained to learn an update rule for few-shot learning. 
To further advance this direction, 
\citet{LeeY2018icml} explicitly split a meta-learner 
into task-specific and task-general components, where each component is updated in a more effective way, compared to update them simultaneously. 

\subsubsection{Set-Input Network~\label{subsec:set}}

Neural networks that are capable of being invariant to permutation and dealing 
with variable-length inputs have recently gained a lot of attention to learn 
semantic representation of sets.
\citet{zaheer2017deep} provide a theoretical justification to 
a unique structure 
of neural networks that is invariant to permutation, which is able to 
being deployed to many interesting applications: multiple instance learning, point-cloud classification, etc.
\citet{EdwardsH2017iclr} develop a generative process of a set 
given a context vector, which is inferred by a statistic network that takes into 
account the exchangeability of a dataset.
Moreover, \citet{LeeJ2018arxiv} propose a feed-forward neural network with self-attention, which still holds permutation-invariant property.

\begin{figure*}[t]
\begin{center}
\includegraphics[width=0.95\linewidth]{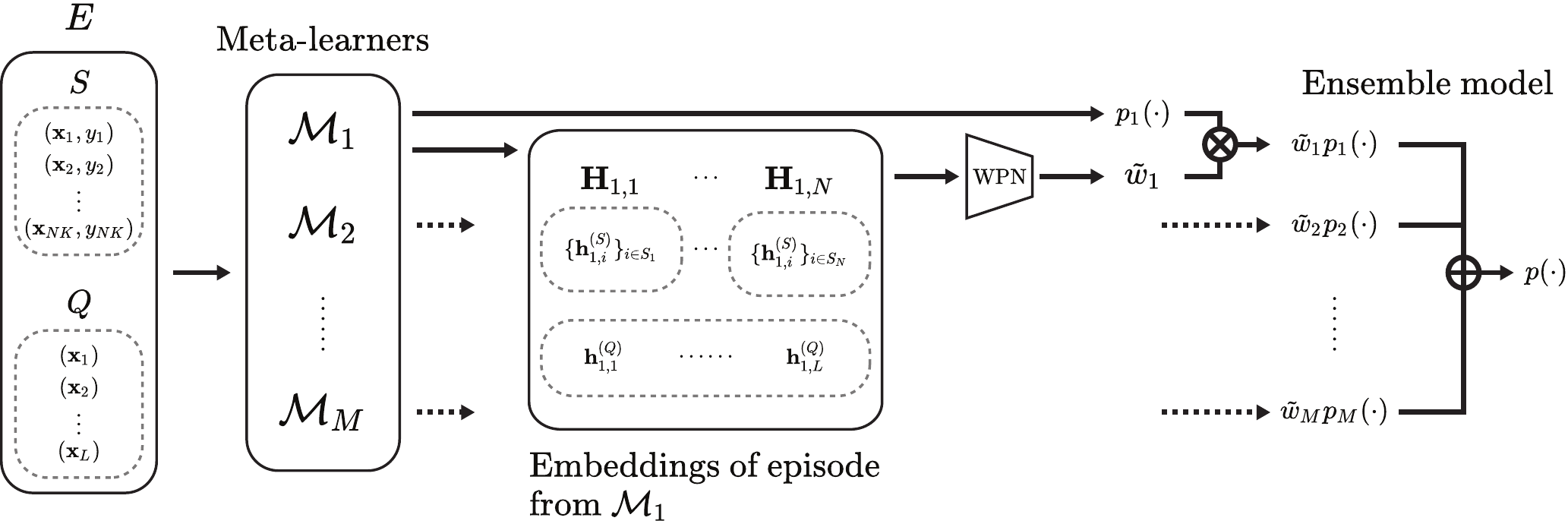}
\end{center}
\caption{Architecture of MxML that produces 
a class probability of the ensemble model.
The mixture coefficient for the base meta-learners
are determined by embedding vectors of episode from each meta-learner.
The inputs of network are a query set $Q$ and a support set $S$.
$m$th meta-learner generates $\bH_{m,1}, \ldots, \bH_{m,N}$ and $\bh_{m,1}^{(Q)}, \ldots, \bh_{m,L}^{(Q)}$,
and the output of MxML is produced by the linear combination of class probabilities of meta-learners with the weights from WPN.
}
\label{fig:mxml}
\end{figure*}

\section{Main Algorithm}

In this section, we introduce our main algorithm
to train mixture of meta-learners in a task-adaptive fashion.

\subsection{Mixture of Meta-Learners (MxML)}

Mixture of meta-learners (MxML) task-adaptively aggregates \emph{base meta-learners}, 
in which mixture coefficients are determined by weight prediction network (WPN). 
Figure~\ref{fig:mxml} introduces an overall structure of MxML
that generates $M$ representations of an episode to 
determine the weight proportion of meta-learners by WPN.
Specifically, an episode composed of $(S,Q)$ is transformed 
by the $m$th base meta-learner as follows:
\begin{equation}
\left(\bh_m^{(S)},\bh_m^{(Q)}
\right) \coloneqq 
\left( 
\{\bh^{(S)}_{m,i}\}_{i = 1}^{NK}, 
\{\bh_{m,i}^{(Q)}\}_{i=1}^L 
\right),
\label{eqn:defrep}
\end{equation}
where $\bh^{(S)}_{m,i}$, $\bh_{m,i}^{(Q)}$ are the hidden representations
of the $i$th instance in the support and query set, respectively.
To ease exploiting the label information, 
we collect $K$ instances that belong to the same label
and denote the hidden representations of them as 
$\bH_{m,n} = \{ \bh_{m,j}^{(S)} \}_{j \in S_n}$,
where $S_n$ means the subset of support set 
that only contains the data labeled as $n$. 
Hence, we denote $\bH_m=\{\bH_{m,n}\}_{n=1}^N$.

Then, the final prediction of MxML 
is established by combining the predictions of meta-learners
with mixing coefficients as follows:
\begin{equation}
p(\bx | S ; \btheta, \bpsi) = \sigma \left(
\sum_{m=1}^M w\left(
\bH_m, \bh_m^{(Q)}; \btheta
\right)
p_m(\bx | S ; \bpsi)
\right)
\label{eqn:mxml},
\end{equation}
where $w(
\bH_m, \bh_m^{(Q)}; \btheta
)$ means the importance of each meta-learner from WPN parameterized by $\btheta$,
$p_m(\bx | S ; \bpsi)$ represents the prediction 
of $\bx \in Q$ by the $m$th base meta-learner, and $\sigma(\cdot)$ is a softmax function.
Details on WPN are described in the subsequent section.

MxML requires a two-step training procedure for base learners and WPN,
in which $M$ datasets are given. For the first step, 
each meta-learner is trained from its associated dataset,
and fixed throughout the next step. MxML allows us to choose any type of
meta-learners including prototypical network~\citep{SnellJ2017nips} and MAML~\citep{FinnC2017icml}.
For the second step, WPN is trained by a series of episodes sampled 
from $M$ diverse datasets. 
In this step, $\btheta$ is trained by minimizing the cross-entropy
between weighted prediction \eqref{eqn:mxml} and the associated labels, 
while $\bpsi$ is fixed after training base meta-learners.
The objective function is introduced as follows:
\begin{equation}
\argmax_{\btheta}
\bbE_{R \sim D}\left[ \bbE_{(S, Q) \sim R} \left[ 
\frac{1}{L} \sum_{(\bx, y) \in Q} y \log p(\bx | S; \btheta, \bpsi)
\right]\right],
\label{eqn:obj}
\end{equation}
where $D$ refers to a dataset distribution which is defined 
in the space that all of the datasets exists, 
$R$ means a single dataset sampled from $D$, 
and $E$ represents an episode from selected dataset $D$. 

\begin{figure}[t]
\begin{center}
\includegraphics[width=0.65\linewidth]{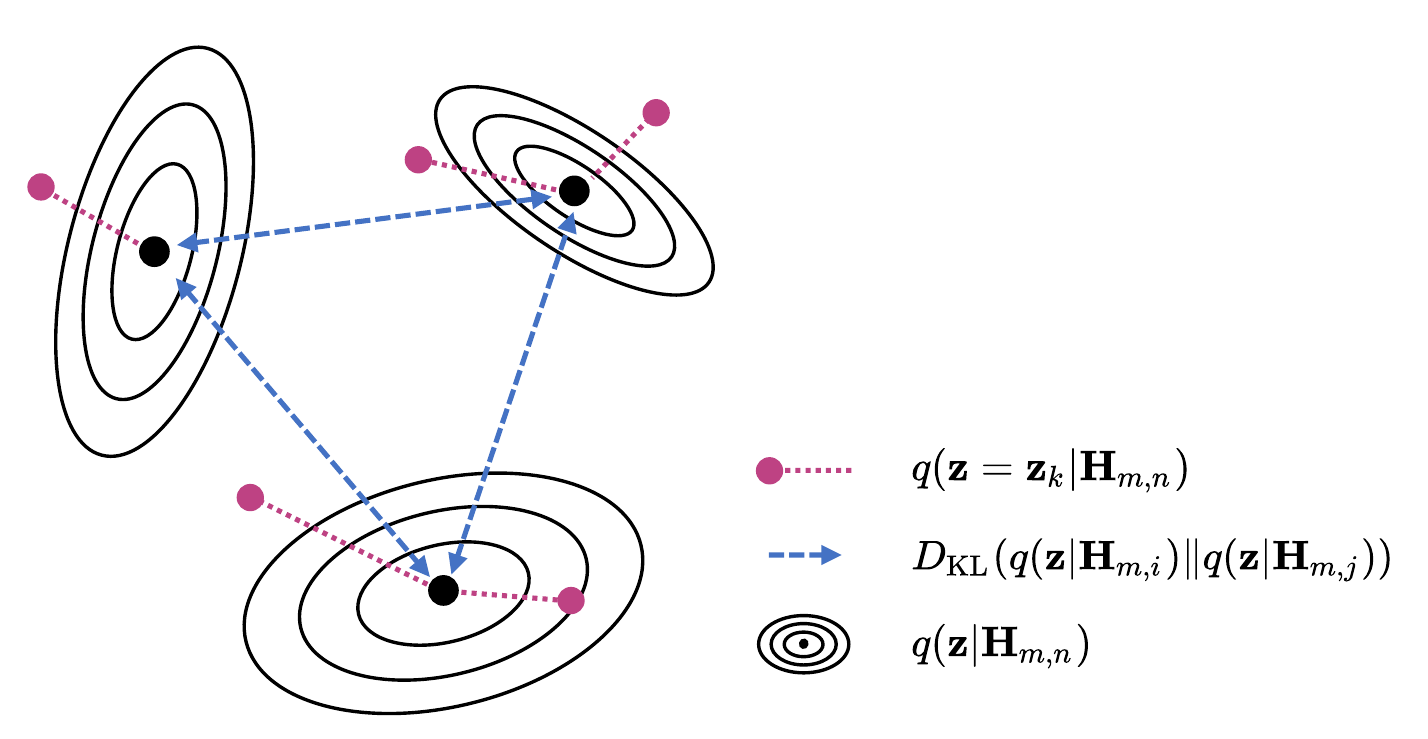}
\end{center}
\caption{
Visualization of components for weight prediction of $m$th meta-learner. 
$q(\bz|\bH_{m,n})$ is a distribution of $n$th class in the support set
(blue circles),
$q(\bz_k|\bH_{m,n})$ is a probability density of a query from 
the most closest distribution (red dot), 
and the KL divergences are denoted as green arrows. 
Predicted weight for the base meta-learner is large if 
the sum of KL divergences is large (classes are separable), 
and product of the log density of query predictions is 
large (queries are correctly predicted with high confidence). 
}
\label{fig:wpn}
\end{figure}

\subsection{Weight Prediction Network (WPN)}

To measure the importance of $m$th learner, 
WPN introduces a distribution that encodes $\bH_{m,n}$ into 
a vector, denoted as $q(\bz | \bH_{m,n}; \btheta)$, which is parameterized by $\btheta$.
We expect that the base meta-learner performs reasonably well when the inter-class 
distributions are separable and the predicted class of a query is highly 
concentrated on a specific class. In this sense, the weight prediction
on $m$th learner is defined as follows:
\begin{align}
w\left(
\bH_m, \bh_m^{(Q)}; \btheta
\right) 
&= \frac{1}{N^2}\sum_{i,j=1}^N D_{\textrm{KL}}
\left(
q(\bz|\bH_{m,i}; \btheta ) \| q(\bz|\bH_{m,j}; \btheta)
\right) \nonumber\\
&\quad+ \lambda\sum_{k=1}^{L}\log \left(S\left(\left\{q\left(\bz = \bz_k| 
\bH_{m,n} ; \btheta \right)\right\}_{n=1}^N
\right)\right)
\label{eqn:wpn}
\end{align}
where $q(\bz| \bH_{m,n} )$ is a distribution over 
the latent vectors that belong to $n$th class in the support set 
(referred to as class-specific distribution),
$\bz_k$ refers the latent vector of $k$th query,
and $S(\cdot)$ is the smooth max function that 
returns the maximum component in the set in a differentiable way. 
We assume that $q(\bz|\cdot)$ is a multivariate normal 
distribution with zero off-diagonal entries.
Mean and diagonal variances of $q(\bz | \bH_{m,i})$ are 
given from the neural network with input 
$\bH_{m,i}$. Since $\bH_{m,i}$ is a set of embeddings, 
set encoding network is needed. 
Bi-directional LSTM is used in \citep{VinyalsO2016nips}
and average pooling-based set representation method is used in 
\citep{zaheer2017deep,EdwardsH2017iclr}.
Likewise, we also use 
average vectors for encoding $\bH_{m,i}$
because of its simplicity. 
Exact implementation details are shown in Table~\ref{detailed_implement}.

The first part of the right-hand side of \eqref{eqn:wpn} explains the distance 
between class-specific distributions.
We assume that the value of the term will increase when the classes 
in the task are more separable than other meta-learners. 
In the second part of \eqref{eqn:wpn}, $\bz_k$ is obtained from the neural network 
with input $\bh_k^{(S)}$. The probability density in the point 
$\bz_k$ from the nearest distribution is multiplied over the entire 
query. Apparently, this term is cumulative, so 
utilizing more number of queries make $w(\cdot, \cdot; \btheta)$ more dependent on 
the second part. Figure~\ref{fig:wpn} shows graphical visualization of each component.

WPN is trained by optimizing the parameters thereof 
from minimizing the cross entropy loss between 
prediction of MxML and the true label (see Algorithm~\ref{alg:wpn}).
This formulation explicitly consider training WPN from  
diverse datasets to generalize on the novel one. 

Our insight to use the parameterized WPN 
and train them to predict the model performance 
is because the evaluation for the model from 
\eqref{eqn:wpn} is not a perfect metric. 
Actually, it is not easy to validate the model 
without any similar instances from the target. 
Since the target task is given as an episode,
we can barely estimate based on the task embedding 
structure. 
By building a structured weight inference 
instead of using the model that inputs a set of 
embedding vectors and outputs a single weight 
prediction, it keeps from over-fitting to a simple
selection of single meta-learner.

\begin{algorithm}[t]
\caption{Optimization of WPN}
\label{alg:wpn}
\begin{algorithmic}[1]
\REQUIRE Base meta-learners learned from each dataset 
$\{ p_{m}(\cdot) \}_{m=1}^{M}$ 
and multiple datasets $\{D_g\}_{g=1}^G$.
\ENSURE Learned parameters $\btheta$.
\STATE Initialize $\btheta$ randomly.
\WHILE {not done}
	\STATE Select $R$ from $\{D_g\}_{g = 1}^G$ randomly.
	\STATE Sample an episode $E \sim R$, where ${E = (S, Q)}$.
	\FOR {$i = 1, 2, \ldots, M$}
		\STATE Determine $w_i$ by \eqref{eqn:wpn}
	\ENDFOR
	\STATE Compute $p(\bx)$ with $\{ w_j \}_{j = 1}^M$ in \eqref{eqn:mxml}.
	\STATE Optimize WPN parameterized by $\btheta$ by minimizing \eqref{eqn:obj}.
\ENDWHILE
\end{algorithmic}
\end{algorithm}

\subsection{Discussion}

In order to handle a novel few-shot learning task, 
the model needs to be trained over similar set of tasks. 
But in case that there is no available similar set of tasks, 
or more specifically, if there is no other training classes 
containing plenty amount of instances 
in the same visual domain with the target, 
then the model should be trained from diverse domain to 
generalize to a novel one. We mainly focus on this problem setting 
which we consider more realistic.

In the ensemble methods in classical supervised learning, 
the base learners should be accurate as possible.
They are evaluated with validating examples which has 
the same distributional property with the test examples, 
and only good ones of them are used to combine (otherwise,
the model degenerates). It is also similar in 
meta-learning phase if we assume that the target tasks 
are achieved from known domain and similar (but not exact) 
tasks can be collected so that we can validate the models, 
then we can possibly attain good ones.

In case of the target task is given from the first seen 
distribution, however, then there are no available 
validating examples for the base meta-learners since it 
contains only few-shot training examples ($S$) with some test instances ($Q$).
Our main proposal is to build a model that evaluates the meta-learner
given an episode to select the best performing model 
depending on distributional property of the task so 
that the model can be capable of solving any kind of tasks.
Since the episode is composed of two sets $(S,Q)$, we 
formulate this as a set-based problem.
Sometimes $S$ contains too small instances to evaluate the model, 
we found that utilizing query instances helps a lot
(i.e., transductive setting).

\begin{table}[t]
\centering
\caption{Description of datasets used in meta-training and -test, which includes 
the number of images $N_{\textrm{img}}$, the number of total classes $N_{\textrm{cls}}$, the number of classes 
used for training meta-learners and WPN, 
and the average image size of each dataset.
Our experiment protocol requires 
splitting classes of each dataset in meta-training into two exclusive sets:
one is used for training base-learners and the other is for training WPN.}
\label{tab:dataset_desc}
\begin{tabular}{llcccc}
\toprule
\textbf{Type} & \textbf{Dataset} & $N_{\textrm{img}}$ & $N_{\textrm{cls}}$ & \textbf{Split} & \textbf{Average of (H, W)} \\
\midrule
\multirow{5}{*}{Meta-train} 
& CIFAR-100 & 60,000 & 100 & (80/20) & (32, 32) \\
& VOC2012 & 11,540 & 20 & (20/-) & (386, 470) \\
& AwA2 & 37,322 & 50 & (40/10) & (713, 908) \\
& Caltech256 & 30,607 & 256 & (205/51) & (325, 371) \\
& Omniglot & 32,460 & 1,623 & (1,298/325) & (105, 105) \\
\midrule
\multirow{5}{*}{Meta-test}
& MNIST & 70,000 & 10 & (-/-) & (28, 28) \\
& CIFAR-10 & 60,000 & 10 & (-/-) & (32, 32) \\
& CUB200 & 11,788 & 200 & (-/-) & (386, 467) \\
& Caltech101 & 9,146 & 101 & (-/-) & (244, 301) \\
& miniImageNet & 60,000 & 100 & (-/-) & (469, 387) \\
\bottomrule
\end{tabular}
\end{table}

\begin{table}[t]
\centering
\caption{Details of 
implementation including input/output 
and layers of components of MxML
for $m$th meta-learner and WPN. 
ConvBlock is composed of series of convolutional layer, 
batch normalization, and ReLU activation function. 
Each convolutional layer contains the kernel with size (3,3)
and the channel with 64 for protoypical network 
and 32 for MAML, as implemented originally. 
Avg refers the mean vector of all instances 
in $\bH_{m,n}$ (see equation \ref{eqn:defrep})
which is fed to dense layer to produce 
$\bmu_{m,n}, \bsigma_{m,n} \in \bbR^{d_z} $, 
means and log diagonal variances  of 
class-specific distribution. 
}
\label{detailed_implement}
\begin{tabular}{lccc}
\toprule
& \textbf{Meta-learner} & \multicolumn{2}{c}{\textbf{WPN}} \\
\midrule
Input & $(S,Q)$ & $ \bH_m $ & $ \bh_m^{(Q)} $ \\
\multirow{2}{*}{Layer} & \multirow{2}{*}{4 ConvBlocks} & Avg, $\textrm{Dense}(2d_z)$ & $\textrm{Dense}(d_z)$ \\
& & $\xrightarrow{} \{(\bmu_{m,n}, \log\bsigma_{m,n}^2)\}_{n=1}^N$ & $\xrightarrow{}\{\bz_k\}_{k=1}^L$ \\
Output & $( \bH_m, \bh_m^{(Q)} )$ & \multicolumn{2}{c}{$w_m$} \\
\bottomrule
\end{tabular}
\end{table}

\section{Experiments\label{sec:exp}}

In this section, we show that our methods outperform other existing methods.
First, we introduce the datasets used in this paper and the detailed settings of the experiments.

\subsection{Datasets\label{subsec:datasets}}

We use various image datasets to train and test our network.
Five datasets are used to train base meta-learners and WPN,
and distinct five datasets are used for the phase of meta-test.
For the fair comparison, we collect datasets in a basis of 
four categories:

\begin{itemize}
\item Gray-scale low/high resolution images.
We use Omniglot~\citep{Lake2015science} 
and MNIST~\citep{lecun2010dataset} in this category.
Omiglot is used for training base-learners and MxML
and MNIST is used for the meta-test phase, because 
the number of classes is not large enough for training meta-learners.
\item Colored low-resolution images.
We use CIFAR-10 and CIFAR-100~\citep{Krizhevsky2009tech},
where the former is used for the meta-test phase
and the latter is used for training meta-learners.
\item Colored high-resolution animal images.
We employ AwA2~\citep{Xian2017tpami} and CUB200~\citep{Welinder2010tech}:
the former consists of high-resolution animal images crawled from web sites
and the latter is composed of images from 50 species of birds.
CUB200 is originally designed for fine-grained classification,
then this is relatively difficult to achieve the high performance.
\item Colored high-resolution generic object images.
We use VOC2012~\citep{Everingham2010ijcv},
miniImageNet~\citep{FeiFei2004cvpr}, and Caltech101/Caltech256~\citep{Griffin2007tech}
in this category. All of them consist of high-resolution images of generic objects.
\end{itemize}

Table~\ref{tab:dataset_desc} summarizes statistics and experiment 
settings of each dataset: which dataset is used for the phase 
of meta-training or meta-test, 
and the number of classes used for training 
meta-learners and WPN. 
We simply remark that all datasets are pre-processed 
in the same fashion, where images are resized into $84 \times 84$ 
resolution and gray-scale images are converted into 3-channel images.

\subsection{Experimental Details\label{subsec:exp_details}}

We set parameters of WPN as $d_z = 128$, $\lambda = 10^{-1}$, 
with fixed learning rate $10^{-4}$ using Adam optimizer~\citep{KingmaDK2015iclr}. 
While training and testing, 15 queries with single meta-batch is given. 
The parameters of prototypical network is set similar to original setting
except for the learning rate that starts from $10^{-3}$ and 
decreases to $10^{-4}$ when it approaches 70 epoch out of 100 epochs. 
We use second order MAML with inner loop learning rate $3 \times 10^{-2}$, 
and the same learning rate used in prototypical network. While training, 
meta-batch size is fixed to 2 in the entire out-of-distribution task 
and 1-shot in-distribution task, and 4 in 5-shot in-distribution task.

In addition, we empirically observe 
that normalized features  
are effective to train WPN.
Thus, in this paper all of the features extracted from base meta-learners 
are normalized as:
\begin{equation}
	\hbh_p = \frac{\bh_p}{\|\bh_p\|_2} \quad \textrm{and} \quad 
	\hbh_q = \frac{\bh_q}{\|\bh_q\|_2},
\end{equation}
where $\bh_p \in \bbR^{d}$ and $\bh_q \in \bbR^{d}$ 
are $d$-dimensional representation of support and query vectors, respectively.
We assume that the normalization helps to standardize a scale in 
embedding spaces learned from different datasets.

\begin{table}[t]
\centering
\caption{
10-way 5-shot classification results of 
MxML with prototypical networks as base meta-learners and baselines.
Transductive and non-transductive setting are denoted as Trans./Non-trans., respectively.
CUB and mImgNet indicate CUB200 and miniImageNet.
The results are obtained 
from average accuracy of 600 episodes with query size 15. 
}
\label{tab:tenfive_result_proto}
\begin{tabular}{llccccc}
\toprule
\multirow{2}{*}{\textbf{Model}}
& \multirow{2}{*}{\textbf{Meta-train}} & \multicolumn{5}{c}{\textbf{Meta-test}} \\
& & MNIST & CUB & CIFAR-10 & Caltech101 & mImgNet \\
\midrule
\multirow{5}{2.5cm}{Dataset-specific\\model} & AwA2 & 75.21 & 38.14 & 27.22 & 55.96 & 32.02 \\
& CIFAR-100 & 77.63 & 29.83 & 45.87 & 61.19 & 35.38 \\
& Omniglot & 73.13 & 17.12 & 17.25 & 29.46 & 21.06 \\
& VOC2012 & 67.24 & 24.72 & 24.42 & 46.65 & 28.71 \\
& Caltech256 & 79.48 & 39.08 & 33.73 & 78.13 & 43.80 \\
\midrule
Single model & \multirow{4}{1.5cm}{Multiple\\datasets} & 76.72 & 39.20 & 39.34 & 75.05 & 44.66 \\
Uniform averaging & & 80.04 & 41.38 &39.46 & 73.88 & 45.10 \\
\textbf{MxML} (Non-trans.) & & 80.25 & 42.62 &43.92 & 77.32 & 46.34 \\
\textbf{MxML} (Trans.) & & \textbf{80.42} & \textbf{43.13} & \textbf{46.17} & \textbf{78.55} & \textbf{46.62} \\
\bottomrule
\end{tabular}
\end{table}

\begin{table}[t]
\centering
\caption{
10-way  5-shot  classification  results  of  MxML  with  MAML 
as  base  meta-learners  and  baselines. 
It follows same settings in Table~\ref{tab:tenfive_result_proto}.
}
\label{tab:tenfive_result_maml}
\begin{tabular}{llccccc}
\toprule
\multirow{2}{*}{\textbf{Model}}
& \multirow{2}{*}{\textbf{Meta-train}} & \multicolumn{5}{c}{\textbf{Meta-test}} \\
& & MNIST & CUB & CIFAR-10 & Caltech101 & mImgNet \\
\midrule
\multirow{5}{2.5cm}{Dataset-specific\\model} & AwA2 & 50.98 & 37.13 & 19.71 & 46.44 & 31.80 \\
& CIFAR-100 & 60.73 & 32.82 & 39.27 & 56.51 & 39.76 \\
& Omniglot & 68.51 & 18.00 & 19.62 & 29.69 & 21.64 \\
& VOC2012 & 48.24 & 22.67 & 19.91 & 38.58 & 28.09 \\
& Caltech256 & 56.56 & 38.49 & 29.69 & 67.27 & 41.16 \\
\midrule
Single model & \multirow{4}{1.5cm}{Multiple\\datasets} & 58.62 & 39.25 & 35.97 & 63.75 & 43.83 \\
Uniform averaging & & \textbf{72.73} & 39.28 & 32.66 & 65.57 & 43.33 \\
\textbf{MxML} (Non-trans.) & & 72.26 & 42.02 & 36.24 & 65.91 & 43.28 \\
\textbf{MxML} (Trans.) & & 72.53 & \textbf{43.73} & \textbf{39.42} & \textbf{67.66} & \textbf{45.73} \\
\bottomrule
\end{tabular}
\end{table}

\subsection{Out-of-distribution}

We design out-of-distribution task to verify that our model
performs robustly on the few-shot classification task sampled 
from first seen distribution. 
We use 5 datasets to train the model (i.e., AwA2, CIFAR-100, Omniglot, VOC2012, and Caltech256), 
and evaluate it from 5 separate datasets (i.e., MNIST, CUB-200, CIFAR-10, Caltech101, and miniImageNet).
The classes of each meta-train dataset are randomly split into 2 subsets. 
One (80\% of entire classes) is used to train class-specific meta-learners, 
and the other (20\% of entire classes) is used to train WPN that learns from 
diverse task distribution.

The results of the experiments are shown in 
Table~\ref{tab:tenfive_result_proto} (with base meta-learner as
prototypical network) and Table~\ref{tab:tenfive_result_maml} 
(with base meta-learner as MAML). 
Dataset-specific models are only trained within their 
associated dataset, and their classification results 
for each meta-test datasets are listed. 
When some datasets share the common source (i.e., the same visual domain 
so they have the similar image resolution) or 
share the similar tasks (classifying among animals, or general objects)
then the class-specific models tend to perform well on the target dataset.
However, that is not always true on some cases such as Omniglot to MNIST. 
We can expect that the model trained on Omniglot performs the most 
on MNIST dataset because both of them consider gray-scale 
character images, but Caltech256 trained meta-learner 
is the stronger classifier.  
This is a supporting reason to building MxML rather than 
using a similar dataset manually because it is difficult to 
notice that model would perform the best in advance. 

Single model has an identical structure with a dataset-specific model, 
but is trained on multiple datasets. In many cases, we observe that 
the performance degrades than well-performing single learner
because of many irrelevant tasks from unrelated domains. 
Uniform averaging model is a mixture model of dataset-specific models 
with identical mixture coefficient. Also in this case, some models 
are degrading.

\begin{figure}[t]
\centering
\includegraphics[width=0.5\linewidth]{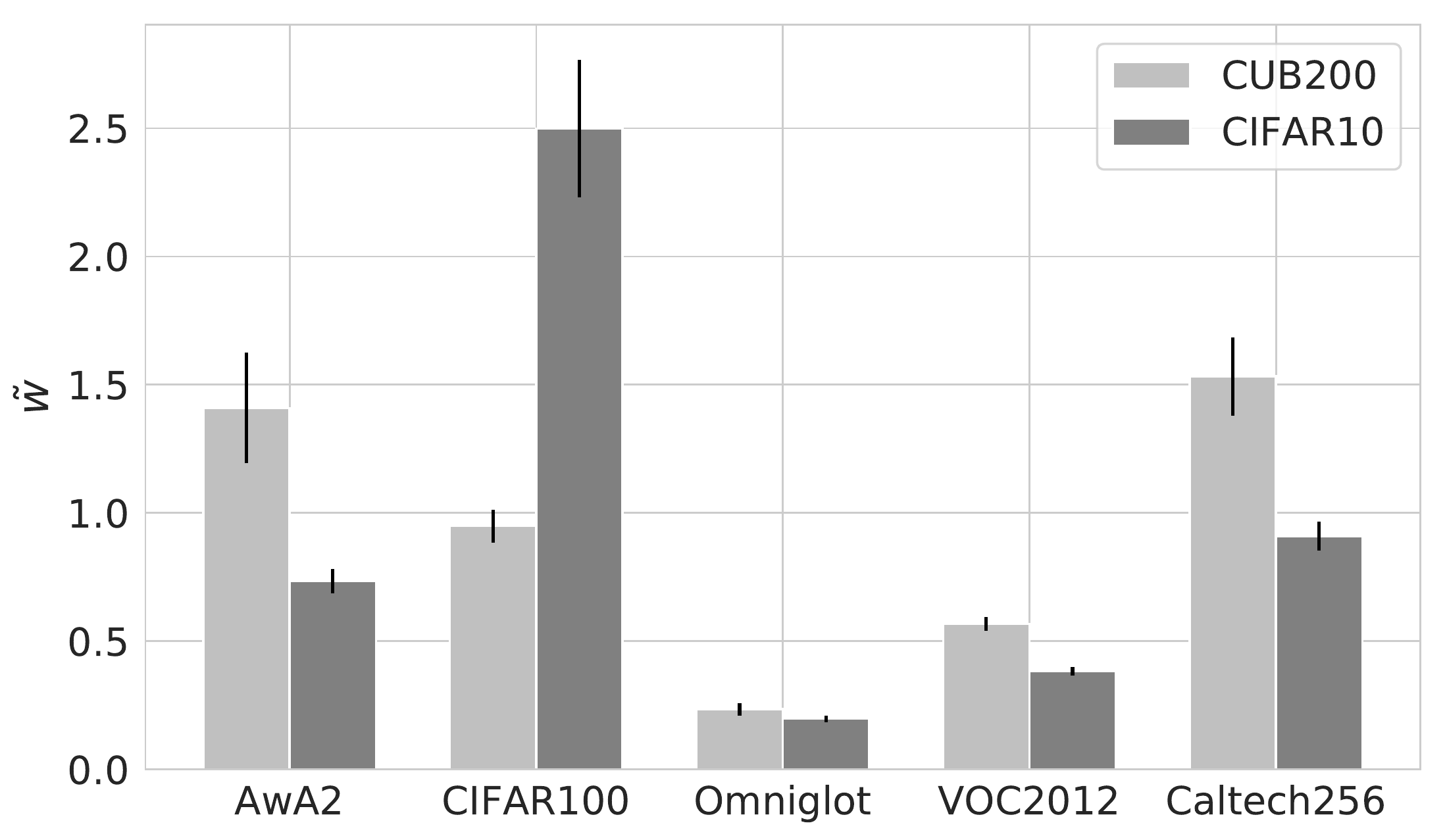}
\caption{Mixture coefficients (as vertical axis) 
on base meta-learners trained from the associated datasets (as horizontal axis) 
for 10-way 5-shot MxML (Trans.). The mean and standard deviation of ensemble weights are computed by 600 episodes from meta-test set.}
\label{fig:average_w}
\end{figure}

Figure~\ref{fig:average_w} shows the averages and standard deviations
of mixing coefficient on base learners associated with datasets, 
when CUB200 or CIFAR10 is given as a meta-test dataset.
When there exists a relevant dataset to the target task, 
MxML assigns relatively high coefficient to the dataset. 
We observe that the weight on the base learner from CIFAR100
is much higher than the ones from other datasets when the 
tasks are given from CIFAR10. On the contrary, MxML assigns 
large coefficient on AwA2 and Caltech256 when the tasks are 
generated from CUB200, where all of them contain the tasks of classifying animals.

\begin{figure}[t]
\begin{center}
\includegraphics[width=0.5\linewidth]{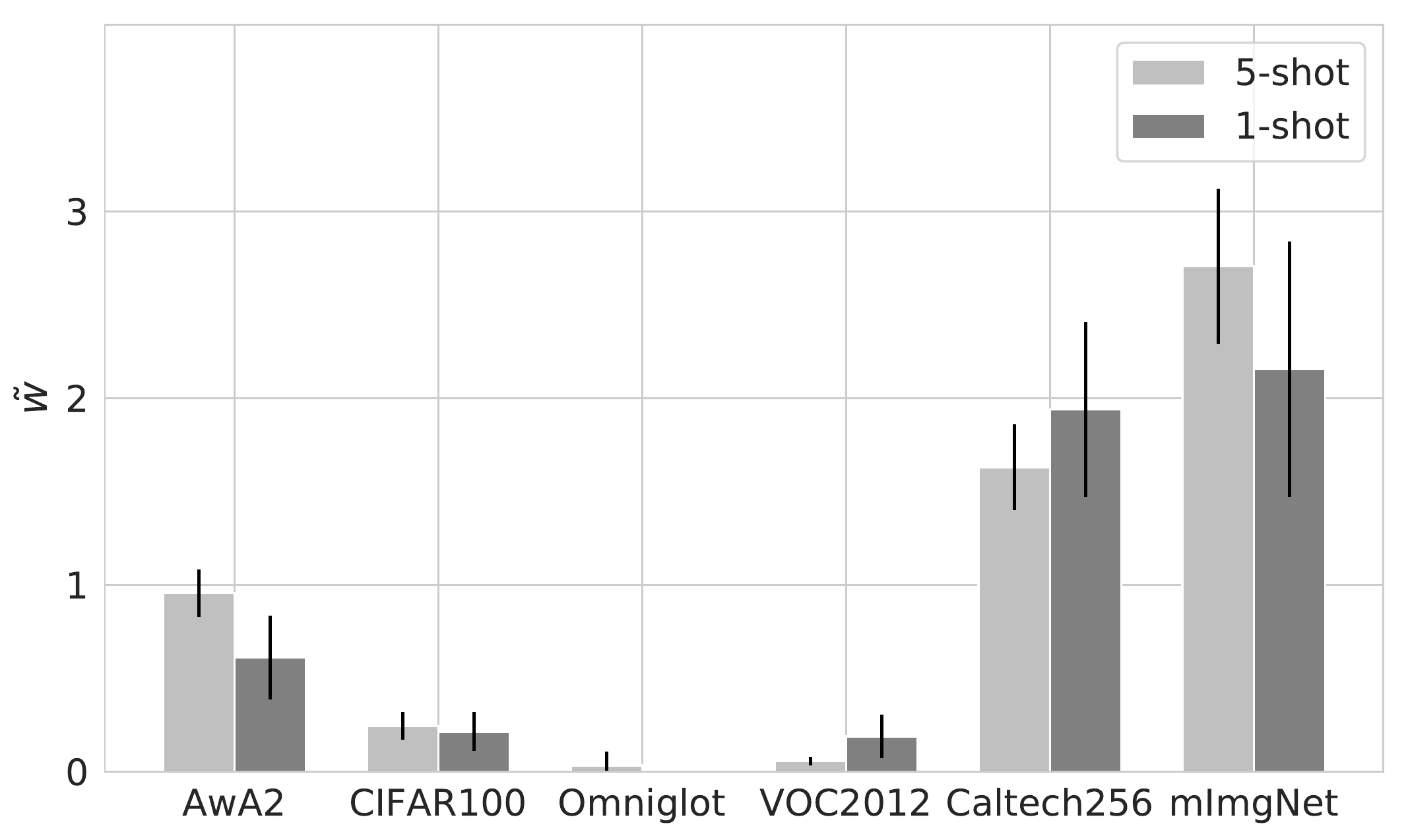}
\end{center}
\caption{Mixture coefficients (as vertical axis) 
on base meta-learners trained from the associated dataset (as horizontal axis), in case of 5-way few-shot classification task from miniImageNet. 
}
\label{fig:avg_weight2}
\end{figure}

\subsection{In-distribution}

In in-distribution task, we show that our model 
performs good as well when the task is sampled 
from the already-seen distribution.
We make slight change from the out-of-distribution experiment setting.
From the setting in Table~\ref{tab:dataset_desc},
miniImageNet (denoted as mImgNet) dataset is split into 
3 subsets as in ~\cite{VinyalsO2016nips} with 
the number of classes 64/16/20 (meta-train/validation/test). 
One base meta-learner trained on miniImageNet dataset 
is added compare to the previous setting, and is tested
only on the miniImageNet dataset.
The result shows that our model consistently 
improves the performance in the conventional 
$N$-way $K$-shot classification problem with additional 
meta-learners trained on other datasets (Table~\ref{tab:mini5result}). 
Figure~\ref{fig:avg_weight2} shows the mixing 
coefficients of base meta-learners,
in which MxML puts more 
attentions on miniImageNet and Caltech256
in both 1-shot and 5-shot experiments.

\begin{table}
\centering
\caption{Accuracy with  95\% confidence interval 
of MxML with prototypical network as a base meta-learners
in 5-way classification task sampled from miniImageNet of which non-overlapping classes are included in the meta-training.}
\label{tab:mini5result}
\begin{tabular}{lcc}
\toprule
\textbf{Model} & \textbf{1-shot} & \textbf{5-shot} \\
\midrule
Dataset-specific model & 50.741 (0.764) & 67.781 (0.664) \\
Single model & 47.432 (0.750) & 65.729 (0.641) \\
Uniform Averaging & 44.618 (0.756) & 66.007 (0.667) \\
MxML & \textbf{51.393}  (0.765) & \textbf{69.338} (0.642)\\
\bottomrule
\end{tabular}
\end{table}

\section{Conclusion\label{sec:conclusion}}

In this paper, we propose a task-adaptive 
mixture of meta-learners, referred to as MxML 
for few-shot classification.
We observe that a common practice for meta-learning has a major limitation:
tasks used in the meta-training and meta-test phases are sampled from the similar task distribution.
To resolve this critical issue, 
we tackle a challenging problem in which a test task is sampled from a novel dataset.
We then propose an ensemble network that
learns how to adaptively aggregate base meta-learners for the given task. 
Extensive experiments on diverse datasets confirm that MxML outperforms other baselines.

\bibliographystyle{abbrvnat}
\bibliography{sjc}

\end{document}